\documentclass[twoside,11pt]{article}

\usepackage{times}
\usepackage{fullpage}

\usepackage{amsmath,amsfonts}
\usepackage{amssymb,amsopn}
\usepackage{bm} 

\newcommand{\eat}[1]{}

\usepackage{color}
\usepackage{graphicx}
\usepackage{subfigure}
\usepackage{multirow}
\usepackage{times}
\usepackage{epsfig}

\usepackage[T1]{fontenc}
\usepackage[font=small,labelfont=bf,tableposition=top]{caption}

\DeclareCaptionLabelFormat{andtable}{#1~#2  \&  \tablename~\thetable}

\begin{document}

\newcommand{\KL}[1]{{\color{blue}Kuan:#1}}
\newcommand{\SX}[1]{{\color{brown}Xing:#1}}
\newcommand{\AK}[1]{{\color{red}Anoop:#1}}
\newcommand{\PN}[1]{{\color{red}#1}}

\newcommand{\thmf}{\texttt{THMF}}
\newcommand{\hmf}{\texttt{HMF}}
\newcommand{\snew}{score$_{new}$}
\newcommand{\sall}{score$_{all}$}
\newcommand{\trank}{\texttt{TRank}}
\newcommand{\rand}{\texttt{Rand}}
\newcommand{\tsort}{\texttt{TSort}}
\newcommand{\lstm}{\texttt{LSTM}}

\title{Temporal Learning and Sequence Modeling for a Job Recommender System}

\author{
Kuan Liu, 
Xing Shi, 
Anoop Kumar,
Linhong Zhu, Prem Natarajan\\ ~~Information Sciences Institute, University of Southern California\\
\{liukuan,xingshi,anoopk,linhong,pnataraj\}@isi.edu
}

\date{}

\maketitle

\begin{abstract}
We present our solution to the job recommendation task for \textit{RecSys Challenge 2016}. The main contribution of our work is to combine temporal learning with sequence modeling to capture complex user-item activity patterns to improve job recommendations. First, we propose a time-based ranking model applied to historical observations and a hybrid matrix factorization over time re-weighted interactions. Second, we exploit sequence properties in user-items activities and develop a RNN-based recommendation model. Our solution achieved $5^{th}$ place in the challenge among more than 100 participants.
Notably, the strong performance of our RNN approach shows a promising new direction in employing sequence modeling for recommendation systems.

\end{abstract}

\section{Introduction}
The problem of matching job seekers to postings~\cite{malinowski2006matching} has attracted lots of attention from both academia and industry (e.g., Xing\footnote{https://www.xing.com/} and LinkedIn) in recent years. \textit{Recsys Challenge 2016} is organized around a particular flavor of this problem. Given the profile of users, job postings (items), and their interaction history on Xing, the goal is to predict a ranked list of items of interest to a user. 

To develop a high quality job recommendation system, one needs to understand and characterize the individual profile and behaviors of users, items, and their interactions. The commonly used factor models \cite{weimer2007maximum,koren2009matrix} learn factors for user and item by decomposing user-item interaction matrices. Neighborhood methods \cite{sarwar2001item,koren2008factorization} rely on similarities between users and items that are derived from content or co-occurrence. These popular methods often ignore or under-exploit important temporal dynamics and sequence properties between users and items.


To address the aforementioned limitations, we explore temporal and sequence modeling to characterize both temporal behaviors and content similarity of users and items. First, we propose a time-based ranking model that leverages  historical interactions for item recommendation. Second, we investigate how to learn latent temporal factors from both user-item interactions and their associated features. Instead of factoring a single aggregated matrix, we extend the context-aware matrix factorization model to explicitly consider temporal interactions.

Finally, motivated by recent success of sequence modeling \cite{graves2009offline, luong2014addressing, sutskever2014sequence, donahue2015long, karpathy2015deep}, we explore the Recurrent Neural Networks (RNNs) approach to capture user-items behavior patterns. We suggest that sequence modeling is very helpful in terms of modeling both item-item similarity and item temporal transition patterns and it consequently leads to a more effective way of utilizing item history. Towards this end, we develop an Encoder-Decoder sequence recommendation system that incorporates feature learning, which significantly increases model flexibility.

The contributions of this work are summarized as follows: 1) A novel temporal ranking approach to 
recommend items from history; 2) An enhanced hybrid matrix factorization model that explicitly incorporating temporal information; 3) A RNN-based sequence model that considerably outperforms matrix factorization models; and 4) Our final system, an ensemble of the above components, achieving $5^{th}$ place in \textit{RecSys Challenge 2016}.

\section{Problem and Data: \textit{RecSys 2016}}
\textit{RecSys Challenge 2016} provides 16 weeks of interactions data  for a subset of users and job items from the social networking and job search website -  \textit{Xing.com}. The task is to predict the items that a set of target users will positively interact with (click, bookmark or reply) in the following week. 

\noindent\textbf{Data Set. }Users and items are described by a rich set of categorical or numerical features or descriptor features. Categorical features take several to dozens of values and descriptor features have a vocabulary size around $100K$. Observation including positive interactions and impressions (items shown to users by \textit{Xing}'s existing recommendation system) at different weeks are also available. The detailed quantitative information of this dataset is shown in Fig.~\ref{fig:data}.

  \begin{figure}[!ht]
    \centering

    \begin{tabular}{|c|c|}
\hline
\textbf{Data splits}             & \textbf{Sizes}\\
\hline
Target/all users       & 150K/1.5M       \\ \hline
Active/all items     & 327K/1.3M      \\ \hline
Interactions & 8.8M \\ \hline
Impressions & 202M \\  \hline 
\end{tabular}
\qquad
\begin{tabular}{|c|c|}
\hline
\textbf{Feature types}  & \textbf{Features} \\ \hline
\multirow{3}{*}{Categorical (U)} & career\_lever, discipline\_id, industry\_id,\\
  &  id, country, region, exp\_years,\\
& exp\_in\_entries\_class, exp\_in\_current\\ \hline
Descriptors (U) & job\_roles, field\_of\_studies \\ \hline
\multirow{2}{*}{Categorical (I)} & id, career\_level, discipline\_id, \\
& country, region, employment \\ \hline
Numerical (I) & latitude, longitude, created\_at \\ \hline
Descriptor (I) & title, tags \\ \hline 
\end{tabular}

    \caption{Statistics of Dataset and Feature Description (U: user; I: item).}
    \label{fig:data}
  \end{figure}


\noindent\textbf{Task and Evaluation Metric. }Given a user, the goal of this challenge is to predict  a ranked list of items from the active item set.
The score is a sum over scores of each user $S(u)$, which is defined as follows:
\begin{align*}
\textbf{S}(u) = &20 * (P@2+P@4+R + \textsc{UserSuccess}) \\
+ &10 * (P@6 + P@20)
\label{eq:score}
\end{align*}
where $P@N$ denotes the precision at $N$, $R$ is the recall, and \textsc{UserSuccess} equals 1 if there is at least one item correctly predicted for that user.


\begin{figure}[!h]
\centering
\includegraphics[width=0.6\columnwidth]{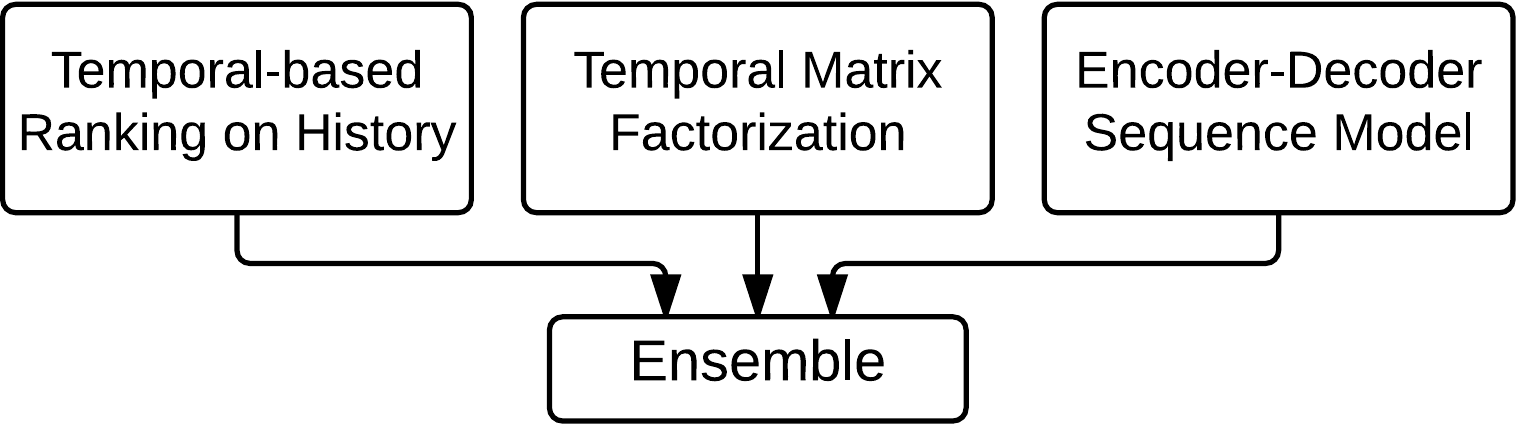}
\caption{System Overview.}\label{pic:overview}
\end{figure}

\section{Methods}
\label{sec:approach}
The proposed solution consists of three main components (see Fig.~\ref{pic:overview}).
We describe each of these in the following.

\subsection{Temporal-based Ranking on History Items}
\label{sec:rank}
The first component of our system is motivated by the observation that \textit{users have a strong tendency to re-interact with items that they already did in the past}. Statistically, on average $2$ out of $7$ items from the first 14 weeks re-appear in the $15^{th}$ week's interaction list. It makes it very plausible to recommend the old items to users. Similarly, items that appeared in the ``impression'' list are also preferred. It motivates us to consider the set of items from past interactions and impressions as a candidate set for user.

Given a user $u$ and an item $i$, the historical interactions between $u$ and $i$ before time $t$ is represented as $M_{i,u,t}\in\mathbb{N}^{K\times T}$, where $T$ is number of time stamps from time 1 to time $t$, $K$ is number of types of interactions (e.g., click, bookmark, or reply), and $M_{i,u,t}(k,\tau)$ is the number of $k$-type interactions at time $\tau\in\{1,\cdots, t\}$.  Given a user $u$, a na\"ive model is to rank each item simply based on the aggregation adoption of history, which leads to 
\begin{equation*}
\small{
S(u, i, t)=\sum_{k}\sum_{\tau=1}^tM_{i,u,t}(k,\tau)
}
\end{equation*}
where $S(u, i, t)$ evaluates how likely user $u$ is to re-interact with an item $i$ given their historical interactions.

However, not every historical interaction by a user has the same importance. For example, a user may prefer re-clicking an item from previous day over one clicked 10 weeks ago.  We conjecture that the importance of user-item interactions depends on the time of interaction. With this assumption, given a user $u$, item $i$ and a particular time $t$, we propose a time reweighed linear ranking model, which is defined as:
\begin{equation*}
S(u, i,t)=wM_{u, i, t}^T
\end{equation*}
where $w$ is the coefficient associated with time, with $w(k,\tau)$ indicating the relative contribution of $k$-type interactions at time $\tau$. 

To learn $w$, we construct triplet constraints $$\mathcal{T} = \left\{u\text{ prefers to re-interacting with } i_1 \text{ to } i_2 \text{ at time } \tau\right\}_{n=1}^N.$$
when \textit{$u$ interacted with $i_1$, $i_2$ before $\tau$, but only interacted with $i_1$ at $\tau$}. We thus obtain the solution of $w$ by minimizing an objective function that incurs a smoothed hinge loss when a constraint is violated. 

\subsection{Factorizing Temporal Interactions}
\label{sec:mf}
\subsubsection{Hybrid Matrix Factorization and Categorical Feature Learning}
\label{sec:hmf}
In order to recommend items that one user interacted with to another similar user or to recommend newly appearing items, we exploit the availability of user/item features, and characterize items/users by vectors of latent factors inferred from their features.

Our approach starts with hybrid matrix factorization technique~\cite{kula_metadata_2015}. To briefly review, we model each user/item as a sum of the representations of its associated features and learn a $d$-dimensional representation for each feature value (together with a $1$-dimensional bias). Let  $\vec{x}^U_j$/$\vec{x}^I_j$ denote the embedding (i.e., vectors of factors) of the user/item feature $j$, $\vec{q}_u$/$\vec{q}_i$ denote the embedding of user $u$ /item $i$, and $b_j^U/b_j^I$ denote the user/item bias for feature $j$. Then
\begin{equation} 
\vec{q}_u = \sum_{j\in f_u} \vec{x}_j^U , \vec{q}_i = \sum_{j\in f_i} \vec{x}_j^I; \qquad b_u = \sum_{j\in f_u} b_j^U , b_i = \sum_{j\in f_i} b_j^I
\end{equation}
The model prediction score for pair \{$u$, $i$\} is then given by
\begin{equation}
\label{eq:score}
S(u,i) = \vec{q}_u \cdot \vec{q}_i + b_u + b_i 
\end{equation}

The model is trained by minimizing the sum of a loss on $S(u,i)$ and the observed ground truth $t(u,i)$, 
\begin{equation}
L = \sum_{\{u,i\} \in I } \ell(S(u,i), t(u,i)) 
\end{equation}
where $I$ is set of interactions between user $u$ and item $i$, $\ell$ is chosen to be Weighted Approximately Ranked Pairwise (WARP) loss~ \cite{usunier2009ranking,weston2010large}, which in our case empirically performs better than other loss functions (e.g. Bayesian Personal Ranking \cite{rendle2009bpr}).

\subsubsection{Temporal re-weighted Matrix Factorization}
\label{sec:thmf}
Our approach is grounded on the assumption that the time factor plays an important role in determining the user's future preference. To this end, we place a non-negative weight associated with time on the loss, which leads to the following equation:
\begin{equation}
L' = \sum_{\{u,i,\tau\} \in I } \ell(S(u,i), t(u,i,\tau)) \times \gamma(\tau) 
\end{equation}
Here the re-weighting term $\gamma$ depends on the time $\tau$ when the user-item interaction happens, which captures the contribution from interactions over time. Additionally, some zero weight $\gamma$ reduces training set size to speed up training and could possibly help prevent over-fitting.

In general $\gamma$ can be learned jointly with other embedding parameters in the model. In practice, we only fixed $\gamma$ as the learned weights $w$ from Model \ref{sec:rank} to speed up training.

\subsection{Sequence Modeling via RNNs}
\label{sec:lstm}

In this section, instead of viewing user-item interactions as independent pairs, we model the entire set of user-item interactions from the same user as a sequence ordered by time. Sequence modeling may reveal the sequential patterns in user-item interactions such as the shifting of user interests over time and the demanding evolving of job items.

\subsubsection{Encoder-Decoder modeling}
\begin{figure*}[!htb]
    \includegraphics[width=15.0cm]{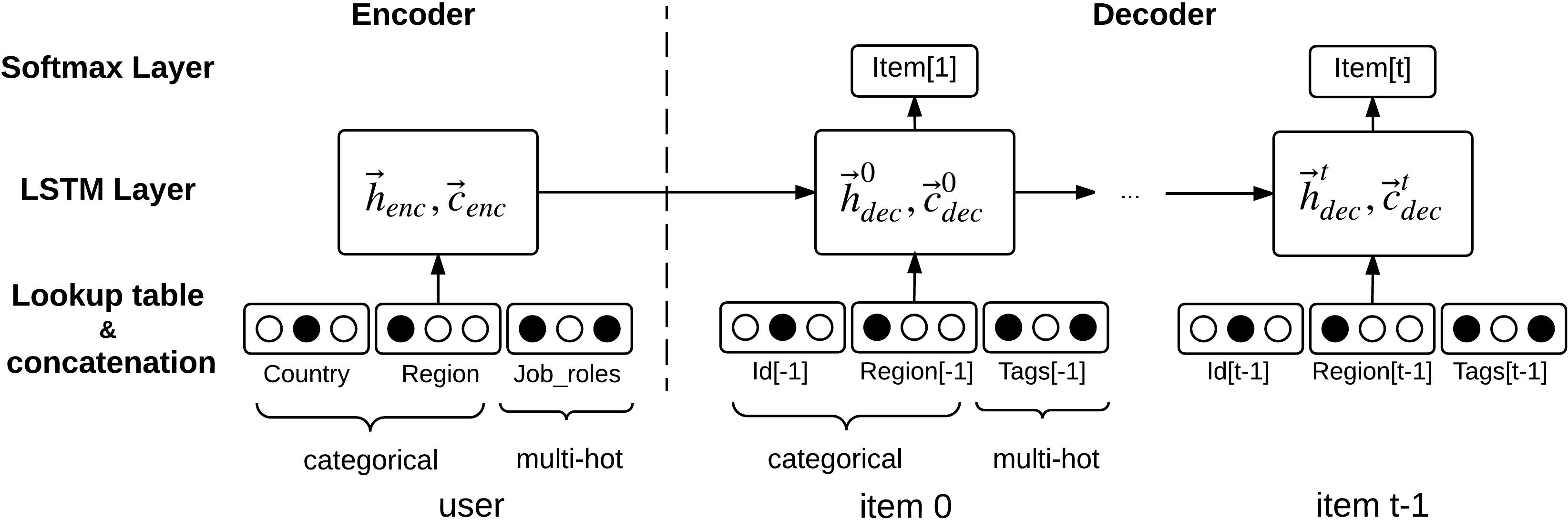}  
  \hspace{0.1em}
\caption{Encoder-Decoder model for recommendation.}\label{pic:rnns}
\end{figure*}

We develop an Encoder-Decoder model \cite{sutskever2014sequence} based on LSTM \cite{hochreiter1997long} 
shown in Fig~\ref{pic:rnns}. Given a user $u$ and its interaction item sequence $\mathtt{I} = \{i^{1}, ... , i^{T}\}$, $u$ is encoded into hidden representation $\vec{h}_{enc}$ and cell state $\vec{c}_{enc}$ via 
    \begin{equation}
    \vec{h}_{enc}, \vec{c}_{enc} = LSTM(f(u), \vec{0}, \vec{0}).
    \label{eq:enc}
    \end{equation}
At decoding phase time step $t$, $\vec{h}_{dec}^{t}$ and $\vec{c}_{dec}^{t}$ are updated by 
\begin{equation}
    \label{eq:dec}
    \vec{h}_{dec}^{t}, \vec{c}_{dec}^{t} = LSTM(f(i^{t-1}), \vec{h}_{dec}^{t-1}, \vec{h}_{dec}^{t-1})
    \end{equation}
and $\vec{h}_{dec}^{t}$ is used to predict $i^{t}$. $i^{0}$ here is a special ``<START>'' item. Cross Entropy is used as the training loss.

\subsubsection{Novel Extensions}

\paragraph{Features.} $f$ in Eq.~(\ref{eq:enc}) (\ref{eq:dec}) and Fig.~\ref{pic:rnns} is a function that maps a user/item index (profile) into a vector by concatenating its features' embedding. For categorical features, the embedding is extracted from a look-up table, and descriptor features are considered as multi-hot features and average pooling is used.
The look-up table is jointly learned during training.
\paragraph{Anonymous users.} Item IDs are used as categorical features to capture item characteristics that are beyond feature descriptions. However, we remove user IDs from user feature set to prevent overfitting to those IDs and empirically observe better performance. It also leads to our natural train/validation set split by randomly splitting user set.
\paragraph{Sampling and data augmentation.} 
Unlike the common success of data augmentation \cite{krizhevsky2012imagenet,simonyan2014very} 
and existing item sampling techniques \cite{improved16Tan}, our results indicate that it is better to use the original sequences, \textit{without sampling items}, to construct training set. Results and analysis are reported in Section \ref{sec:res}. 


\section{Experiments}

\subsection{Settings}
\label{sec:exp_set}
We take user-item interaction data from the $26^{th}$ to the $44^{th}$ weeks as training data and validate our model on the $45^{th}$ week. Submitted results come from models re-trained on data from $26^{th}$ to $45^{th}$ week under the same hyper-parameters. We observe very strong correlation between validation and test scores for all our models and thus \textbf{mostly report validation scores} below due to submission quota.



\subsection{Results}\label{sec:res}
\subsubsection{Recommend from History}
\begin{figure}[t]
\centering
\subfigure[interactions]{\label{fints}\includegraphics[width=0.485\columnwidth]{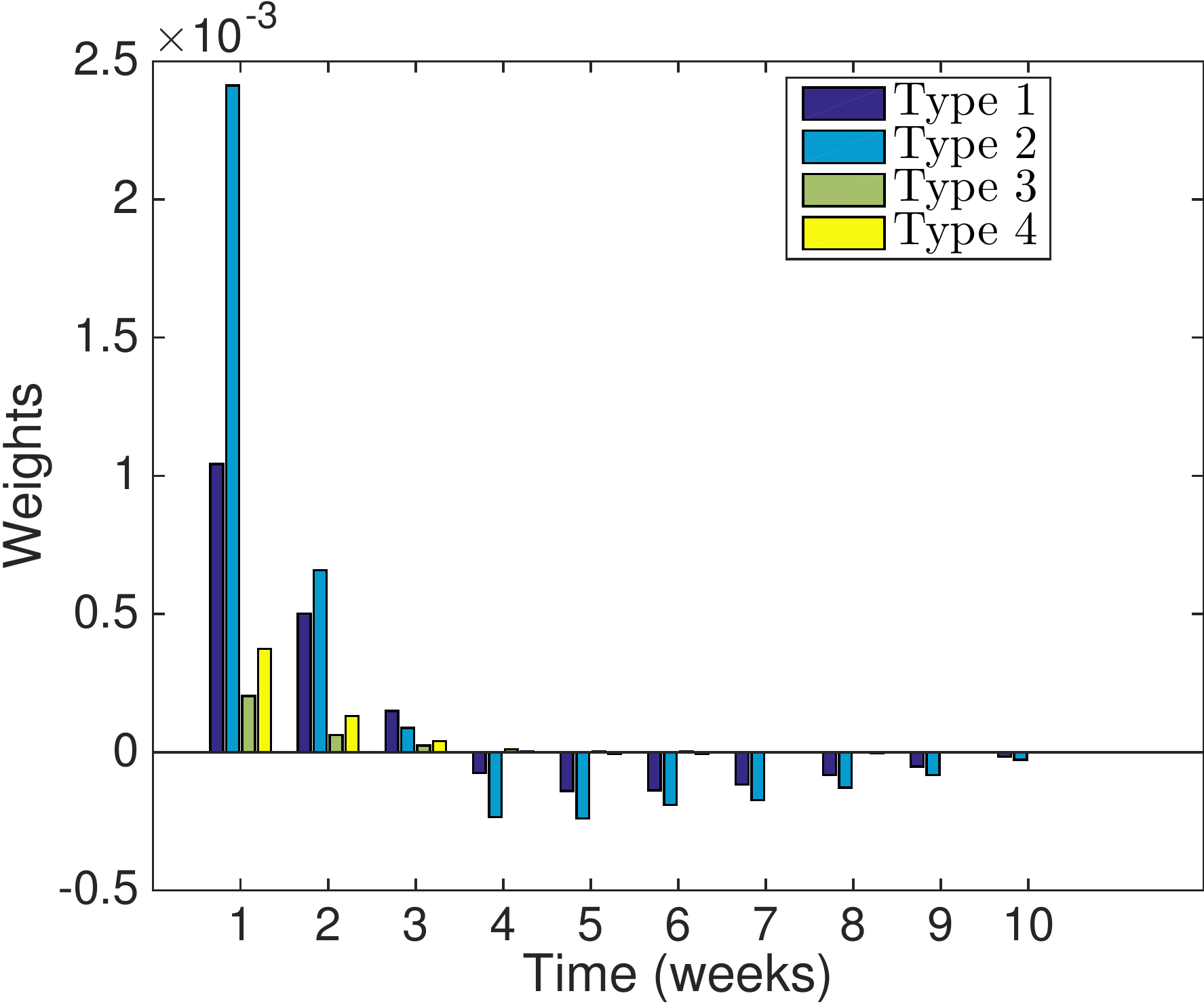}
}
\hspace{-0.2cm}
\subfigure[impressions]{\label{fimps}\includegraphics[width=0.485\columnwidth]{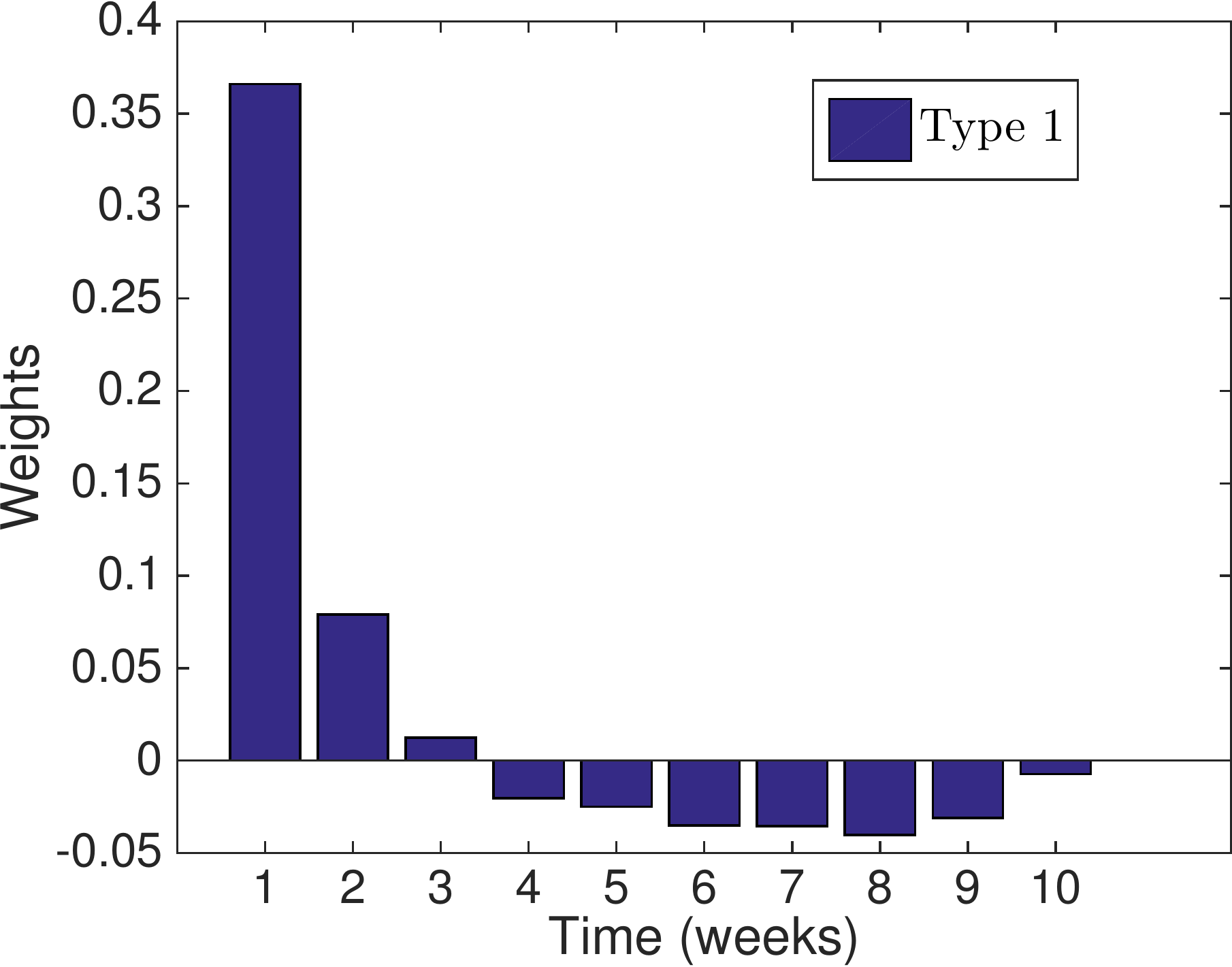}
}
\vskip -0.5em
\caption{Weights learned in Model \ref{sec:rank} for interactions \ref{fints} and impressions \ref{fimps}. $K$=4 for \texttt{INTS} and $K$=1 for \texttt{IMPS}, where type 1 denotes user-item impression pairs, and type 2,3,4 denote click, bookmark and reply, resp. }\label{pic:weights}
\vskip -1.0em
\end{figure}

Our model in Section~\ref{sec:rank} (\trank) is compared to two baseline models: randomized score (\rand) and recency-based sorting (\tsort) that sorts items by the latest time they appear in the history. The results on historical ``interactions''(\texttt{INTS}), ``impressions''(\texttt{IMPS}), and their combinations (\texttt{\texttt{INTS}+IMPS}) are reported in Table~\ref{tHistory}. \trank~clearly outperforms the other two in all the three cases. 
Figure~\ref{pic:weights} shows the learned weights $w$ associated with the designed temporal features. The coefficients are decaying with time in both Figure~\ref{fints} and~\ref{fimps} across different types of interactions, indicating that more recent interactions have a larger statistical impact over the users' future preferences. Furthermore, although recency is important, simply using the latest time performs worse than \trank, which smoothly combines the most recent interactions with historical interactions using the learned weights $w$.

\begin{table}
\small
\centering
\caption{Scores in thousands ($\mathtt{K}$) based on history interactions.}
\label{tHistory}
\vspace{0.5em}
\begin{tabular}{cccc}\hline
Models            & \rand& \tsort & \trank    \\ \hline
\texttt{INTS} & 266       &       284& \textbf{299}      \\ \hline
\texttt{IMPS} &  324      & 375       & \textbf{380}\\ \hline
\texttt{INTS+IMPS} &  463       & 509       & \textbf{524} \\ \hline
\end{tabular}
\vskip -1.0em
\end{table}

\subsubsection{Recommend via Matrix Factorization}
We compare hybrid matrix factorization model in \ref{sec:hmf} (\hmf) and the model in \ref{sec:thmf} (\thmf) with different number of latent factor $d$, without and with features. Two important measures are used: \sall~and \snew. \sall~is the challenge score and \snew~is the score after removing all history user-item pairs. We found \snew~more important in model ensemble and chose in our experiments to early stop model training at the best \snew.

As shown in Table \ref{tMF}, \thmf~models achieve significant improvements on \sall~and \snew~for all $d$, with and without features. Meanwhile, the time comparison shows that the best models achieved by \thmf~require significantly less training time. 

\begin{table}
\centering
\caption{Scores ($\mathtt{K}$) achieved by hybrid matrix factorization models and training time (in hours $h$) .}
\label{tMF}
\begin{tabular}{|c|c|c|c|c|c|c|c|}
\hline
\multicolumn{2}{|c|}{Models} & \multicolumn{3}{c|}{\hmf} & \multicolumn{3}{c|}{\thmf} \\ \hline
Fea& $d$   & \sall                & \snew         &T      & \sall               & \snew        &T       \\ \hline
\multirow{4}{*}{No}&16  & 235                 & 61     &   8.8           & 269                & 65           &   2.8     \\ 
\cline{2-8}
&32  & 301                   & 71      &  3.4           & 320                & 75          &   1.5      \\ 
\cline{2-8}
&48  & 313                 & 78   &    7.7            & 326                & 84                 & 1.7  \\ 
\cline{2-8}
&64 & 330                 & 76    &  3.3             & 340                & 86                & 0.7  \\ 
\hline
\multirow{3}{*}{Yes}&16 & 311                  & 124    &    74           & 361                & 146            &  34     \\ 
\cline{2-8}
&32 & 326                   & 125    &    26           & \textbf{381}               &  \textbf{148}       & 14           \\ 
\cline{2-8}
&48 & 354                 & 128       &      76      & 378                & 144            &   12    \\ \hline
\end{tabular}
\vspace{-1.0em}
\end{table}

Finally, we use items in the ``impression'' list in last week and treat them as ``interactions'' (with $0.01$ down-weight). This boosts performance as seen in Table \ref{tIMP}.
\begin{table}
\centering
\caption{Scores ($\mathtt{K}$) by \thmf~with some ``impression''s as additional observation inputs.}
\label{tIMP}
\begin{tabular}{ccc}\hline
Observations            & INTS & INTS + IMPS    \\ \hline
\sall &  381       &       \textbf{438}     \\ \hline
\snew &  148   &       \textbf{164}     \\ \hline
\end{tabular}
\vskip -1.0em
\end{table}                                                                                                                                                                                                           
\vspace{-1.0 em}
\subsubsection{Recommend via LSTMs}
\noindent\textbf{Setting, Tuning, Details.}
With extensive tuning, we choose to train the encoder-decoder model (\lstm) using a single layer LSTM with hidden vector size as 256 and dropout rate as 0.6. All parameters are uniformly initialized before $[-0.08, 0.08]$. The learning rate of Stochastic Gradient Descent (SGD)
is initially set as 1.0 and will decay with rate 0.7 once the perplexity on development set starts to increase. We only consider the top 50,000 frequent item and replace the remaining item as ``<UNK>''. The final result is ensembled using 6 models with different random seeds.

\noindent\textbf{Performance.} Table \ref{tLSTM12} reports the comparison between \hmf, \thmf~and \lstm. Models are trained on the datasets\footnote{To make fair comparison, all models are trained on full user set and active item set. \hmf, \thmf~do benefit from training with additional items (see Table \ref{tMF}); however, currently we don't have efficient implementation of \lstm~supporting additional items.} with and without features. 
Provided with features, \lstm~obtains superior results to the rest. It verifies that sequence modeling is a promising direction for job recommendation tasks.

We also note when features are not provided, \lstm ~does not show advantages to \thmf. We don't know the exact reason yet but suspect it is due to the inflexibility of non-feature sequence model which has a hard time capturing item transition pattern.

 \begin{minipage}{0.95\textwidth}
  \begin{minipage}[b]{0.52\textwidth}
    \centering
    \small
\begin{tabular}{|c|c|c|c|c|c|c|}
\hline
Fea  &  \multicolumn{3}{c|}{No}  &  \multicolumn{3}{c|}{Yes}  \\ \hline
Models   &  \hmf  & \thmf & \lstm & \hmf & \thmf & \lstm    \\ \hline
\sall  & 313  &  \textbf{347}  &  313  & 312     & 366  & \textbf{391}       \\ \hline
\snew     & 78  & 87    & \textbf{89}    & 104     & 130  & \textbf{140}      \\ \hline
\end{tabular}
\label{tLSTM12}
\vspace{3em}
    \captionof{table}{Scores ($\mathtt{K}$) comparison among \hmf, \thmf, and \lstm~models. All models are trained on active item set.}
  \end{minipage}
  \hfill
  \begin{minipage}[b]{0.40\textwidth}
    \centering
    \epsfig{file=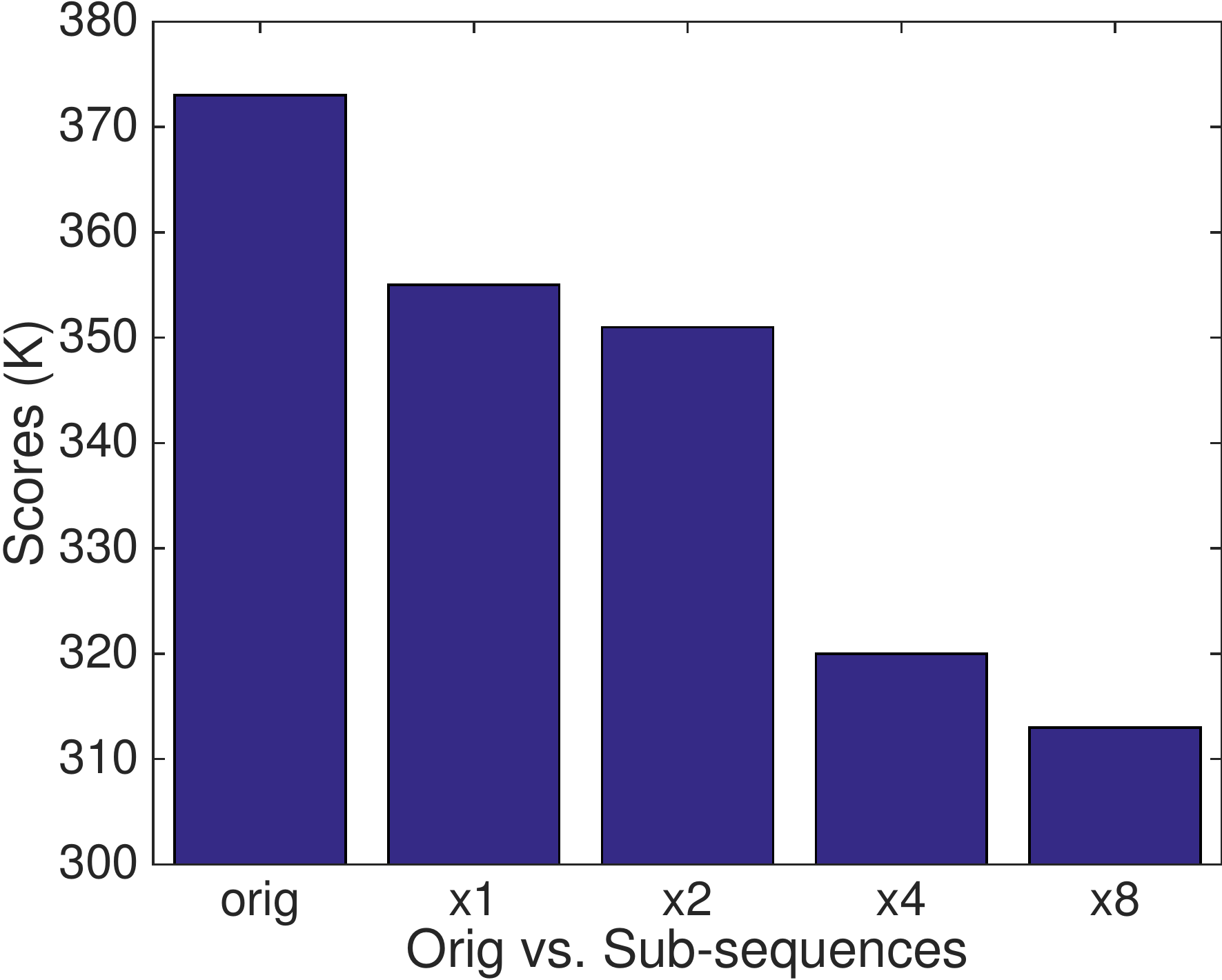, height=1in, width=0.8\columnwidth}
    \label{pic:order}
    \vspace{-0.3em}
      \captionof{figure}{\lstm~scores ($\mathtt{K}$) on original and ``manipulated'' data sets. ``Orig'' denotes the complete sequence. $x_N$ denotes the manipulated data set obtained by randomly sampling sub-sequence proportional to $N$ times.}\label{pic:order}
    \end{minipage}
        \vspace{1.0em}
  \end{minipage}
  
\noindent\textbf{Does sequence help?}
When applying sequence modeling to the recommendation problem, we implicitly assume that sequence or order provides additional information beyond that provided by item frequency alone. To test the validity of this assumption, we generated new training data through sampling sub-sequences in which items were dropped out with certain probability. At the same time, on the average, item appearance frequency would remain unchanged in data with more sampled sub-sequences. 

Experimental results with the new generated training data are shown in Fig \ref{pic:order}. First, increasing sub-sequence sampling leads to decreasing scores (from $x_1$ to $x_8$); Second, original data set (full sequences) gives the best score. These results suggest that item sequences do indeed provide additional information and merit further investigation.




\subsubsection{Ensemble and Final Results}
Given the prediction and confidence scores from the above model components, we perform model ensemble to obtain our final result. The final scores are achieved by using a Random-Forests-based ensemble approach, which outperforms the linear fusion by greedy grid search in our preliminary experiment.
Final scores for different components and the ensemble model are reported in Table \ref{tFinal}.

\begin{table}
\vspace{-.5em}
\centering
\caption{Final Component and Ensemble Results.}
\label{tFinal}
\vspace{0.5em}
\begin{tabular}{ccccc}\hline
Component & History            & MF (ints+imps) & LSTMs & Ensemble    \\ \hline
Valid & 524 & 438  &  391     & 613    \\ \hline
Test & 502 & 441 & 384  & 615\\ \hline
\end{tabular}
\end{table}


\section{Related Works}
Deep Feedforward Networks have been successfully applied in recommender systems. \cite{salakhutdinov2007restricted} used Restricted Boltzmann Machines for Collaborative Filtering and achieved remarkable results. Other feedforward models (e.g. Convolutional Neural Networks, Stacked Denoising Autoencoders) have also been used to extract feature representations from items to improve recommendation \cite{van2013deep,wang2015collaborative}.

\cite{hidasi2015session}~introduced RNNs to recommendation system on the task of session based recommendation. They devised a GRU based RNNs and demonstrated good performance with one hot encoding item input and rank based loss functions. Further improvements on session based recommendation include exploiting rich features like image \cite{hidasiparallel} and data augmentation \cite{improved16Tan}.


\section{Conclusions}
In this paper, we presented our innovative combination of new and existing recommendation techniques for \textit{RecSys Challenge 2016}. Empirical study verified the effectiveness of 1) utilizing historical information in predicting users' preferences and 2) both temporal learning and sequence modeling in improving recommendation. 

Notably, the proposed RNN-based model outperforms the commonly used matrix factorization models. In the future, we would like to extend our research in model designs (e.g. to incorporate features in the output layer and to support other loss functions) and in result analysis to understand why and when the sequence modeling really helps recommendation.   



\bibliographystyle{apalike}
\bibliography{recsys16}

\end{document}